# Ideal Reformulation of Belief Networks*

**John S. Breese**
Rockwell International Science Center
Palo Alto Laboratory
444 High Street
Palo Alto, CA 94301

**Eric J. Horvitz**
Medical Computer Science Group
Knowledge Systems Laboratory
Stanford University
Stanford, California 94305

## Abstract

The intelligent reformulation or restructuring of a belief network can greatly increase the efficiency of inference. However, time expended for reformulation is not available for performing inference. Thus, under time pressure, there is a tradeoff between the time dedicated to reformulating the network and the time applied to the implementation of a solution. We investigate this partition of resources into time applied to reformulation and time used for inference. We shall describe first general principles for computing the ideal partition of resources under uncertainty. These principles have applicability to a wide variety of problems that can be divided into interdependent phases of problem solving. After, we shall present results of our empirical study of the problem of determining the ideal amount of time to devote to searching for clusters in belief networks. In this work, we acquired and made use of probability distributions that characterize (1) the performance of alternative heuristic search methods for reformulating a network instance into a set of cliques, and (2) the time for executing inference procedures on various belief networks. Given a preference model describing the value of a solution as a function of the delay required for its computation, the system selects an ideal time to devote to reformulation.

*This work was supported by Rockwell International Science Center and the National Science Foundation under Grant IRI-8703710.

## 1 Introduction

For a large class of AI problem-solving techniques, great gains in efficiency can be achieved by expending effort on a preliminary meta-analysis of a problem instance before directly executing a solution. Belief-network algorithms highlight the necessity of reformulating or restructuring problem instances. The reformulation of a belief-network can greatly increase the efficiency of inference. Indeed, many belief-network algorithms rely on some preliminary reformulation procedure. Our analysis of reformulation is motivated by our pursuit of techniques for the dynamic construction and solution of belief networks [7, 1].

To date, investigators have made use of offline analyses for reformulating a small number of networks that will be solved many times. Unfortunately, straightforward offline analyses of reformulation, may not be effective in systems that must construct and solve belief-network problems at run time. The computational effort expended for reformulating a newly constructed belief network is not available for the primary task of performing inference with the network. Thus, in time-dependent decision contexts, there is a tradeoff between the time dedicated to reformulating the network and the time applied to the implementation of a solution.

We shall describe the *metareasoning-partition problem* and present principles for computing the ideal partition of resources under uncertainty for several prototypical classes of uncertainty and utility. In Section 3, we shall consider the global optimization of the apportionment of resources to precursory re-formulation for the situation where the reformulated instance is solved once. In Section 4, we will discuss the inclusion of evidence about the progress of problem solving, in a formulation of the metareasoning-partition problem centering on a myopic optimization policy. Following the presentation of theoretical results, we shall discuss in Section 5 an empirical study of the application of these principles to belief networks. We focus, in particular, on an empirical analysis of the ideal amount of time to devote to searching for clusters in belief networks. We acquire and apply probability distributions that characterize the performance of alternative heuristic search methods for finding cliques and the time for executing inference procedures on various belief networks. Given a preference model, describing the value of a solution as a function of the delay needed for its computation, the system determines the ideal time to devote to reformulation.

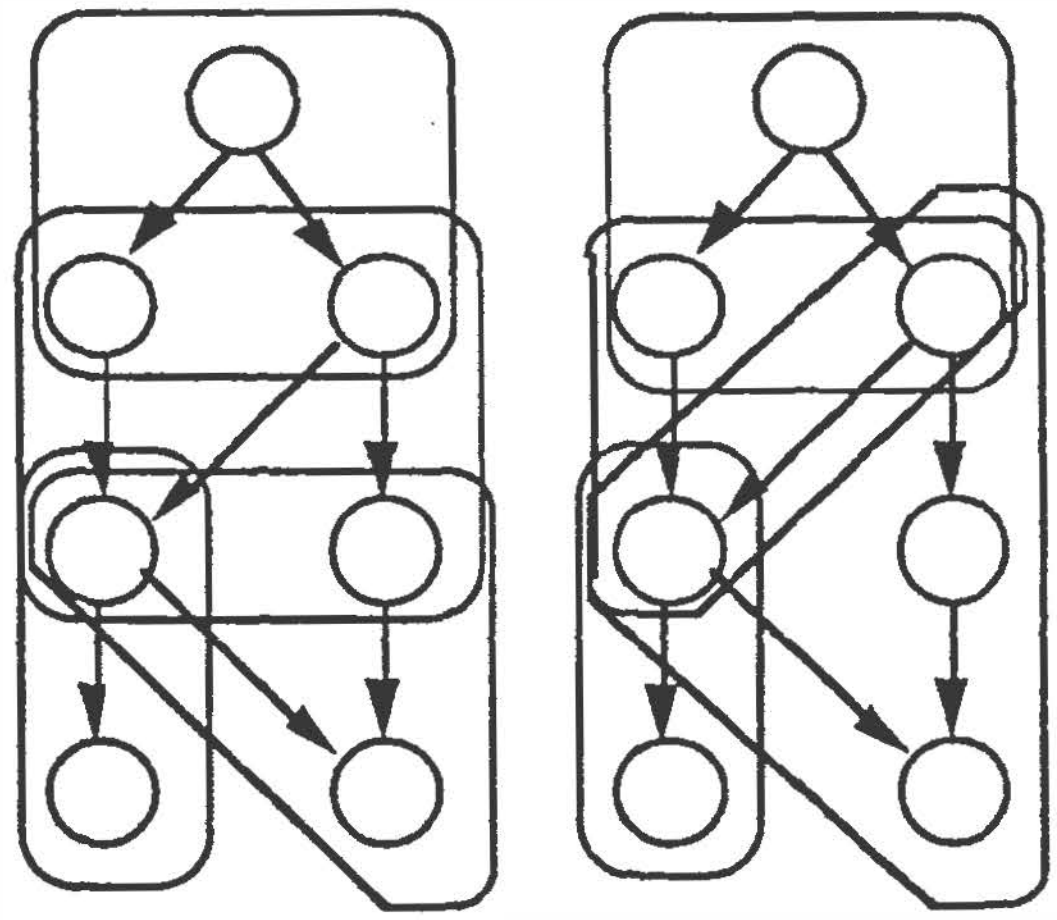

Figure 1: Reformulation of a belief-network instance for the clique-tree approach involves generating and evaluating alternative sets of cliques. Individual cliques are encircled. As highlighted by the graphs, nodes can be members of several cliques.

# 2 Reformulating Belief Networks

Brute-force approaches to the solution of belief-network inference problems are intractable. In a brute-force analysis, we generate a joint distribution by taking the product of all assigned distributions. Given the joint distribution, we compute the marginal probability for any value of a variable or Boolean combination of values, by summing over the relevant dimensions of the joint distribution. The size of the joint distribution is exponential in the number of variables. Thus, although this naive approach is conceptually simple, it requires computation that is exponential in the number of variables.

Although the problem of probabilistic inference with belief networks is $\mathcal{NP}$-hard, methods have been developed for exploiting independence relations to avoid the explicit calculation of the joint-probability distribution. A variety of exact methods has been developed to operate on specific topologies of belief networks [13]. Other recent methods forego exact calculation of probabilities; these approximation techniques produce partial results as distributions or bounds over probabilities of interest [4, 14, 8].

Several promising exact and approximate approaches rely on the reformulation of multiply connected networks. We are studying the ideal control of reformulation of belief network instances with the *clique-tree* approach developed and refined in [11, 9], with Pearl's *method of conditioning* [12], and with the *nested-dissection* method of Cooper [3]. The clique-tree reformulation approach seeks to convert multiply connected belief networks into a corresponding singly connected network of *cliques*. A precursory reformulation of a belief-network instance works to identify cliques, defined as maximal sets of nodes that are completely interconnected. An algorithm has been developed to propagate evidence within this tree of cliques, which is somewhat analogous to the propagation of belief in a singly connected network of variables. Alternative clique-tree reformulations are pictured in Figure 1. For the method of conditioning, reformulation seeks to break loops in a multiply connected belief network, by identifying and instantiating a loop cutset. At solution time, each cutset node must be instantiated with each possible value (or combination of values). Each instance is solved as a separate singly connected belief network problem. Reformulation methods work by generating and evaluating cutsets that minimize the number of problem instances that must be evaluated.

Identifying the best cutset and identifying the best set of cliques are $\mathcal{NP}$-hard problems, since in general they require searching all sets of subsets of the nodes in a belief network. However, we can develop heuristic strategies (see for example [10]) and *flexible* (or *anytime*) search strategies that can deliver increasingly better reformulations as we increase the amount of reformulation time.

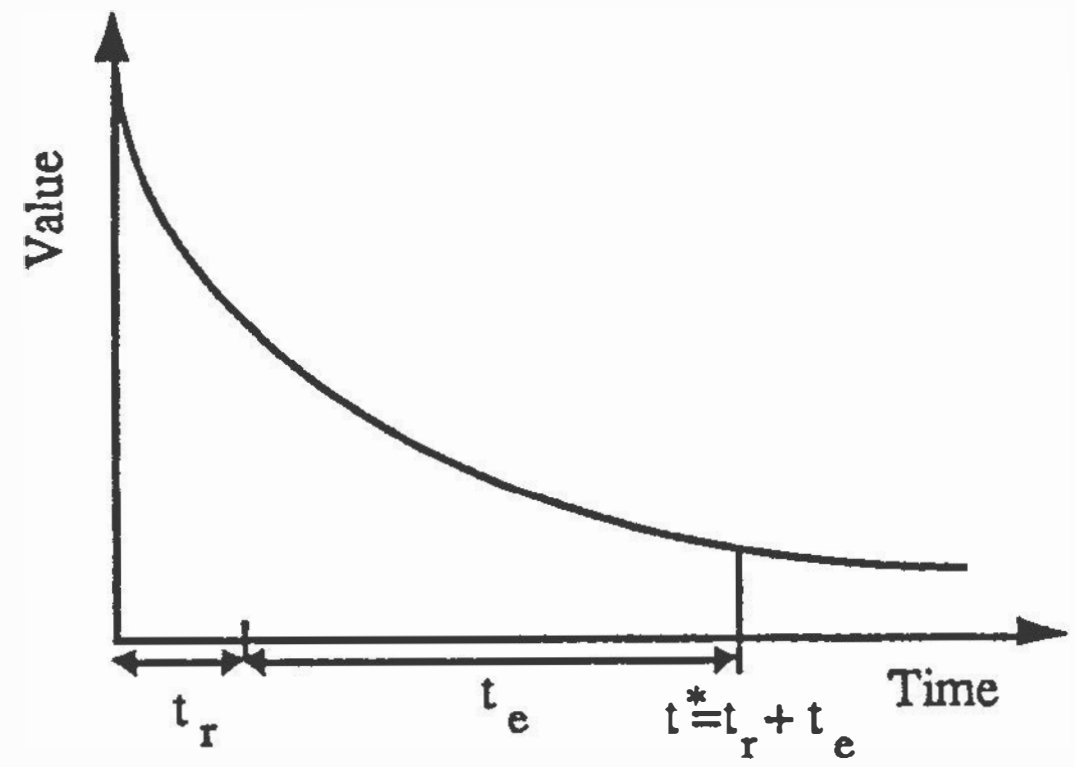

Figure 2: A graph showing the value of a computed result, as a function of the total time required to generate a result. The total delay $t$ is the sum of the time needed for reformulating and solving a belief network, $t = t_r + t_e$.

# 3 Ideal Partition of Resources for Belief Networks

We now shall outline the problem of ideally apportioning resources to the reformulation of belief networks under conditions of uncertainty and describe the application of this problem to the solution of belief networks that are created at run time.

## 3.1 General Problem

We refer to the problem of ideally apportioning resources between a meta-analysis and the solution of a base problem as the *metareasoning-partition* problem [6]. The ideal partition of resources depends on the architecture of an agent, on the availability and form of knowledge and metaknowledge about problem-solving, and on the problem instance at hand. Most meta-analyses for the reformulation of belief networks center on a search process. Thus, we cast reformulation in terms of search. Let $t_r$ be the time the reasoner spends on reformulating a problem instance, and let $t_e$ be the time required to execute a reformulated instance to generate a final solution. Thus, the total time required to solve the problem is $t = t_r + t_e$. Let us assume that the value of a computed result is a function solely of the time at which it becomes available. We express this relationship as $V(t)$, illustrated in Figure 2.

For some inference procedures it may be possible to determine $t_e$ precisely, given the amount of time spent searching for a solution. In this special situation, we can characterize computation time in terms of a deterministic function, $t_e = \mathcal{R}(t_r)$. In this case, the selection of $t_e$ fully determines the time that the solution to the problem becomes available. In general, we must consider the ***uncertainty*** in the relationship between the search time and the time required to compute a desired result.

Under uncertainty, the computation time is characterized by probability distributions for different values of search time,

$$p(t_e|t_r,\xi)$$

where $\xi$ includes any background knowledge about the problem and solution methods (e.g., problem size, hardware parameters, architecture of reasoning system) which may effect these distributions. Our objective is to choose a value of $t_r$ that maximizes the expected value of the computation, given a specification of a value function and distributions for $t_e$. More formally, we seek to maximize the expected value of the result, with respect to $t_r$ as follows:

$$\max_{t_r} \mathrm{E}_{t_e}(V|t_r,\xi) =$$

$$\max_{t_r} \int_{t_e} V(t_r + t_e)p(t_e|t_r,\xi)dt_e \qquad (1)$$

Details of a formal analysis of this problem appear in [2]. Here we highlight the central results for prototypical models of cost.

## 3.2 Deadline Models

The class of ***deadline*** problems captures situations where the cost incurred with delay for a computed result is 0 or insignificant until a deadline $a$ is reached. Thus, an analytical result obtained before time $a$ has value $k$. If the result is not available by time $a$, the result is worthless. We can model a deadline situation via a stepwise value function

$$V(t) = \begin{cases} k & t \leq a \\ 0 & \text{otherwise.} \end{cases}$$

Through substituting this step-function utility into our general formulation (1), we find that the expected value is

$$\begin{aligned} \mathrm{E}_{t_e}(V|t_r,\xi) &= \int_{t_e} V(t_r + t_e)p(t_e|t_r,\xi)dt_e \\ &= \int_0^{a-t_r} kp(t_e|t_r,\xi)dt_e \\ &= kp(t_e \leq a - t_r|t_r,\xi) \qquad (2) \end{aligned}$$

The last term is the probability that the time required for executing the reformulated solution is less than the time remaining after the reformulation process. Thus, for the deadline case, maximizing the expected value is equivalent to maximizing the probability of completing the computation before the deadline.

### 3.2.1 Polynomial Urgency Models

Let us now explore the general model of reformulation under uncertainty where the overall value of a computational result is a polynomial function of the time it becomes available. We consider a model of urgency where the value function $V(t)$ is an $n$th degree polynomial:

$$V(t) = \sum_{i=1}^{n} a_i t^i = \sum_{i=1}^{n} a_i (t_r + t_e)^i$$

where the $a_i$ are constants that are customize the model to particular contexts. Substituting the polynomial form into our general formulation, Equation 1, we seek to maximize:

$$\begin{aligned} & \mathrm{E}_{t_e}(V|t_r,\xi) \\ = & \int_{t_e} \sum_{i=1}^{n} a_i (t_r + t_e)^i p(t_e|t_r,\xi) dt_e \\ = & \int_{t_e} \sum_{i=1}^{n} a_i \sum_{j=0}^{i} \binom{i}{j} t_r^j t_e^{i-j} p(t_e|t_r,\xi) dt_e \\ = & \sum_{i=1}^{n} a_i \sum_{j=0}^{i} \binom{i}{j} t_r^j m^{(i-j)}(t_e|t_r) \end{aligned}$$

where

$$m^{(n)}(x|y) = \int_x x^n p(x|y) dx$$

is the $n$th *moment* of $x$ given $y$.[1]

To maximize the expected value of a computed result defined by Equation 3, we set the derivative of this expression to zero. At the optimum we obtain

$$\sum_{i=1}^{n} a_i \sum_{j=0}^{i} \binom{i}{j} [i t_r^{i-1} m^{(i-j)}(t_e|t_r) + t_r^i \frac{dm^{(i-j)}(t_e|t_r)}{dt_r}] = 0 \quad (3)$$

Thus, for any context of urgency, that can be represented (or approximated) with a polynomial value function of order $n$, we can determine an ideal time to dwell on reformulation $t_r$, given the moments and derivatives of the probability distribution, $p(t_e|t_r,\xi)$ for $i = 1,\ldots,n$. For high-order polynomials, solving Equation 3 and estimating the derivatives may be difficult. However, for linear and quadratic forms, the solution is straightforward.

[1] The first moment of a distribution, $m^{(1)}$, is the expectation.

## 3.3 Target Model

The class of *target* problems refers to situations where the value of a computed result is 0 unless it is *available exactly at time a*. That is, a result obtained before or after time $a$ is worthless. This model is associated with events that must be coordinated tightly under bounded resources, such as time-dependent communications and datasharing over limited bandwidth channels. We can model target problems by representing our value as a delta function

$$V(t) = \delta(a - t) = \delta(a - (t_r + t_e))$$

At a solution for this functional form, we have

$$\frac{dp(a - t_r|t_r)}{dt_r} = 0$$

The derivative of a probability distribution is zero at the mode of a unimodal distribution. This result dictates that, for target models, we should continue searching until the distribution over the expected execution time achieves its mode exactly at time $a$.

# 4 Incremental Analysis of Metareasoning Partition

We have described how we can characterize the efficacy of metareasoning processes for different types of belief networks by acquiring probability density functions about the relationship between $t_r$ and $t_e$ for a large set of instances for each class. Such sampling yields probability distributions $p(t_e|t_r,\xi)$ that we can use to calculate the optimal time to spend on metareasoning. $\xi$ includes any background knowledge about the problem and solution methods which may effect these distributions. So far, we have assumed that we do not have additional knowledge about the problem besides these distributions. If the reasoner is limited to a single-step solution-planning process, where a single meta-analysis is applied to generate an ideal reformulation policy, we are indeed forced to make use of a probability distribution that describes the relationship between $t_r$ and $t_e$ for an entire class of problems. However, we may wish to expend additional resource on an incremental meta-analysis, and make use of detailed information about the relationship between $t_r$ and $t_e$ that is revealed over time.

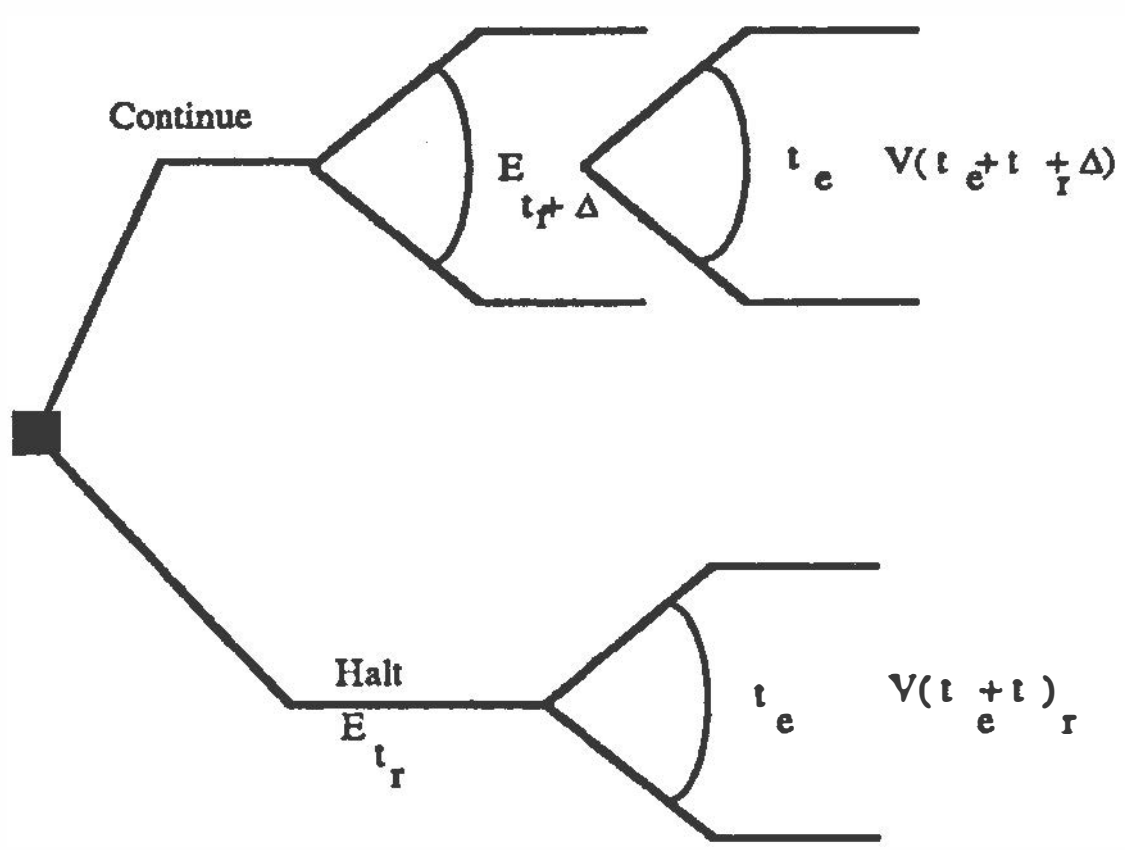

Figure 3: The incremental metareasoning-partition decision problem, incorporating the acquisition and use of information about the progress of reformulation to control the extent of reformulation.

That is, we can assess probability distributions and incorporate information about the *progress of reformulation* as a useful class of evidence for determining the efficacy of future reformulation efforts.

In the incremental approach, we analyze recent reformulation behavior to make a decision about the value of continuing to perform reformulation for an additional, prespecified increment of reformulation time. We can make use of uncertain knowledge of the form

$$p(t_e|t_r, E_{t_r}, \xi)$$

where $E_{t_r}$ refers to evidence observed at time $t_r$ in the progression of reformulation of the current instance. Associated with each time point and evidence $E$ is the actual data structure which embodies the current reformulated problem instance. In general, $E_{t_r}$ can include the complete sequence of evidence or information collected during the process of reformulation.

At each time $t_r$, we must decide whether to halt immediately—and to begin to solve the current problem formulation—or to continue the reformulation search for another $\Delta t_r$. We can express the value of continuing as a lottery over possible results of further reformulation search; we can sample a large set of cases to acquire probability distributions about changes in distributions about $t_e$ as more time is spent on reformulation.

With the incremental metareasoning-partition analysis, we must continually determine if the expected value of the lottery of continuing is greater than the value of halting and solving the current formulation of the belief-network problem. We halt when the expected change in the value of solving the problem after another $\Delta t_r$ is nonpositive. If we decide to continue, we reformulate for another $\Delta t_r$ and again examine a new decision to halt or to continue with a new lottery. The incremental decision tree is displayed in Figure 4. Note that we are not solving for the sequence of choices over time, but rather we develop a greedy, hill-climbing procedure for incremental reformulation. Expected value of halting reformulation:

$$EV_{halt} =$$

$$\int_{t_e} V(t_r + t_e) p(t_e|t_r, E_{t_r}, \xi) dt_e$$

If we continue reformulation, there is a spectrum of evidence $E$ which may be observed.

$$EV_{continue} =$$

$$\int_{t_e} \int_E V(t_r + \Delta t_r + t_e) p(t_e|t_r + \Delta t_r, E_{t_r+\Delta t_r}, \xi)$$

$$p(E_{t_r+\Delta t_r}|E_{t_r}, \xi) dE dt_e \quad (4)$$

The further assumption that $t_e$ is conditionally independent of $t_r$ given the evidence observed so far yields

$$p(t_e|t_r + \Delta t_r, E_{t_r+\Delta t_r}, \xi) = p(t_e|E_{t_r+\Delta t_r}, \xi) \quad (5)$$

simplifying the expression for Equation 4. The criterion for halting is

$$EV_{halt} \geq EV_{continue} \quad (6)$$

The efficacy of the incremental approach relative to an *a priori* reformulation policy depends on the structure of the problem and the costs of performing the incremental analysis [5]. In some cases, an *a priori* analysis can prove that an incremental approach is unnecessary; we show in [2] that for certain value function–distribution pairs, dominance relationships can determine the ideal reformulation policy in advance. When this type of simplification is not possible, there are other factors to consider. If a reasoner cannot obtain access to evidence ($E$) about the time-dependent behavior of a reformulation method, we must treat the method as a mysterious "black box," and base decisions on an *a priori* consideration of summary distributions for large classes of problems. Finally if the cost of evaluating Equation 6 is high, then the overhead of metareasoning may overwhelm potential benefits. In fact, creation of distinctions and models for $E_{t_r}$ and its dynamics that are at once informative and concise are critical to the value of the entire approach.

# 5 Example: Clique Reformulation

We have applied an incremental analysis of ideal partition of resources to the example of clique reformulation of a belief network. The fundamental cycle is construction of a belief network, formation of the clique tree (reformulation), and finally performing inference (calculating the posterior probability of all unobserved variables). We shall present several details about the clique identification strategies. After, we shall describe the procedures used to collect probability distributions for use in the analysis. Finally, we shall review the results of using these distributions in an incremental analysis.

## 5.1 Clique Reformulation Methods

The clique formation methods we examine are based on construction of a *join tree*. The join tree is constructed by the following sequence of steps [16, 13]:

1. Create a Markov network from the original network by interconnecting the parents of each node and removing directionality from the arcs.
2. Calculate an ordering for the nodes.
3. Fill in edges between predecessors of each node in the graph, using the ordering generated in Step 2.
4. Construct the join tree by identifying the *cliques* (subgraphs which are completely connected) in the filled-in graph.

Our analysis of reformulation strategies focuses on Step 2, the generation of an ordering. In particular, we examine a method developed by Kjærulff [10], which we refer to as *K-search*. The better-known procedure for ordering is *maximum cardinality search* (MCS) [16]. The MCS approach starts with an arbitrary node and assigns the next number to the node having the largest set of unnumbered neighbors. K-search generates an ordering by first finding a node whose neighbors form a clique already. If no such node exists, the algorithm uses a cost metric, based on the size of the state space of the neighbors of a node, to determine which node to index next.

## 5.2 A Flexible Clique-Reformulation Strategy

We implemented flexible versions of the MCS and K-search reformulation strategies. The MCS and K-search strategies are both sensitive to the initial ordering of a belief network. The search state space is generated by varying the initial conditions to produce a large number of join tree topologies. After we generate an ordering, we determine the join tree structure. We then estimate the time required to solve that configuration with an efficient estimation procedure [15]. In our study of ideal clique-tree reformulation, we based this time estimate on the sum of the state-space sizes for the cliques in the tree. These "generate and test" procedures maintain a record of the best clique configuration found to that time and continues to search until the procedure is terminated. The strategies are flexible in that they generate solutions that are monotonically increasing in quality (decreasing in $t_e$), and make available, at all times, the best join tree found so far. As the reformulation time, $t_r$, is increased, the procedure searches additional join-tree configurations.

## 5.3 Classes of Data about Reformulation Efficacy

As we discussed in Sections 3 and 4, we need to make use of uncertain knowledge that relates the time needed for execution of a problem formulation to the time spent on generating the reformulation. We have obtained distributions from a frequency analysis of the various reformulation strategies described above. We used IDEAL, a general influence-diagram programming environment, to collect this distributional information [15]. We directed the system to construct random belief networks of different sizes and connectedness, and to apply reformulation algorithms to the networks. We collected data for many networks to generate statistics regarding $p(E_{t_r+\Delta t_r}|E_{t_r},\xi)$ and $p(t_e|E_{t_r},\xi)$.

### 5.3.1 Run-Time Estimate and Reformulation

One useful class of knowledge for making decisions about the partition of resources focuses on estimated run time as a function of reformulation time. As discussed above, increasing the time for reformulation increases the number of cliques that the program has explored and scored, with $E$ defined as the best estimate encountered so far. To investigate the efficacy of the randomized K-search procedure, we generated a large number of networks and collected data about the trajectory of improvement in run-time estimates as additional time was spent on reformulation. Several of these trajectories are displayed in Figure 4. The trajectories are normalized to show the proportional decrease in estimated run time as a function of time. Our analysis revealed that the incremental time used for generating new configurations has only modest impact on total execution time.

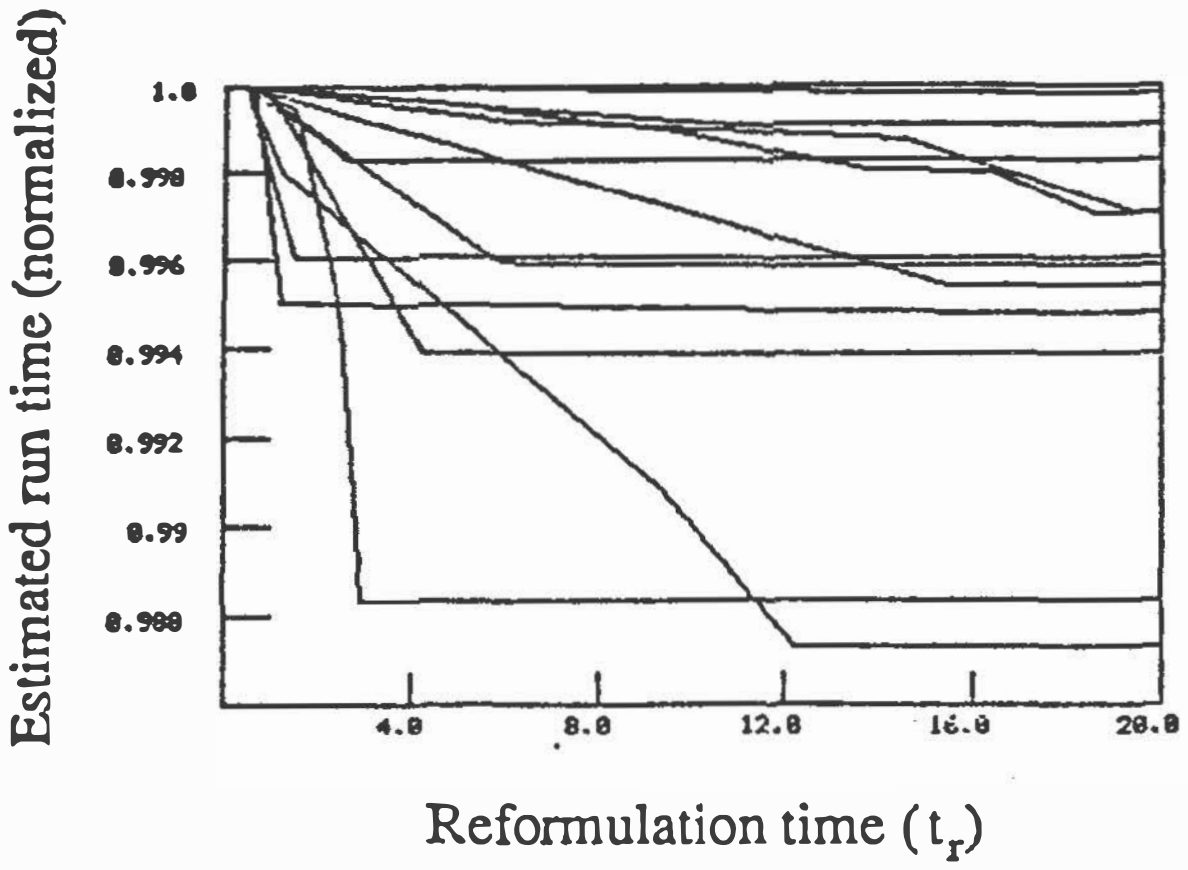

Figure 4: The proportional reduction in run-time estimate as a function of reformulation time for the K-search reformulation for a set of randomly generated networks.

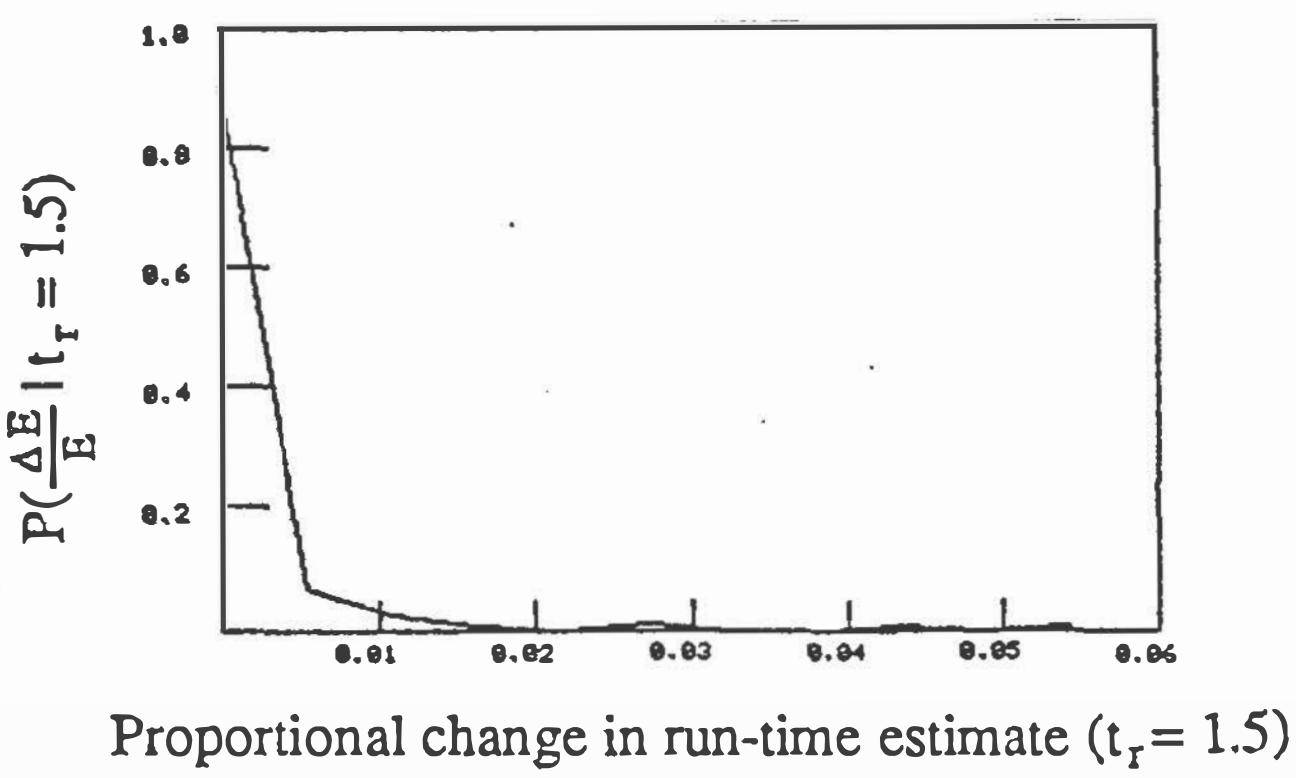

Figure 5: The probability of various levels of decrease in running time for K-search for an increment of reformulation search for $t_r = 1.5$, generated from a sample of 200 randomly generated belief networks.

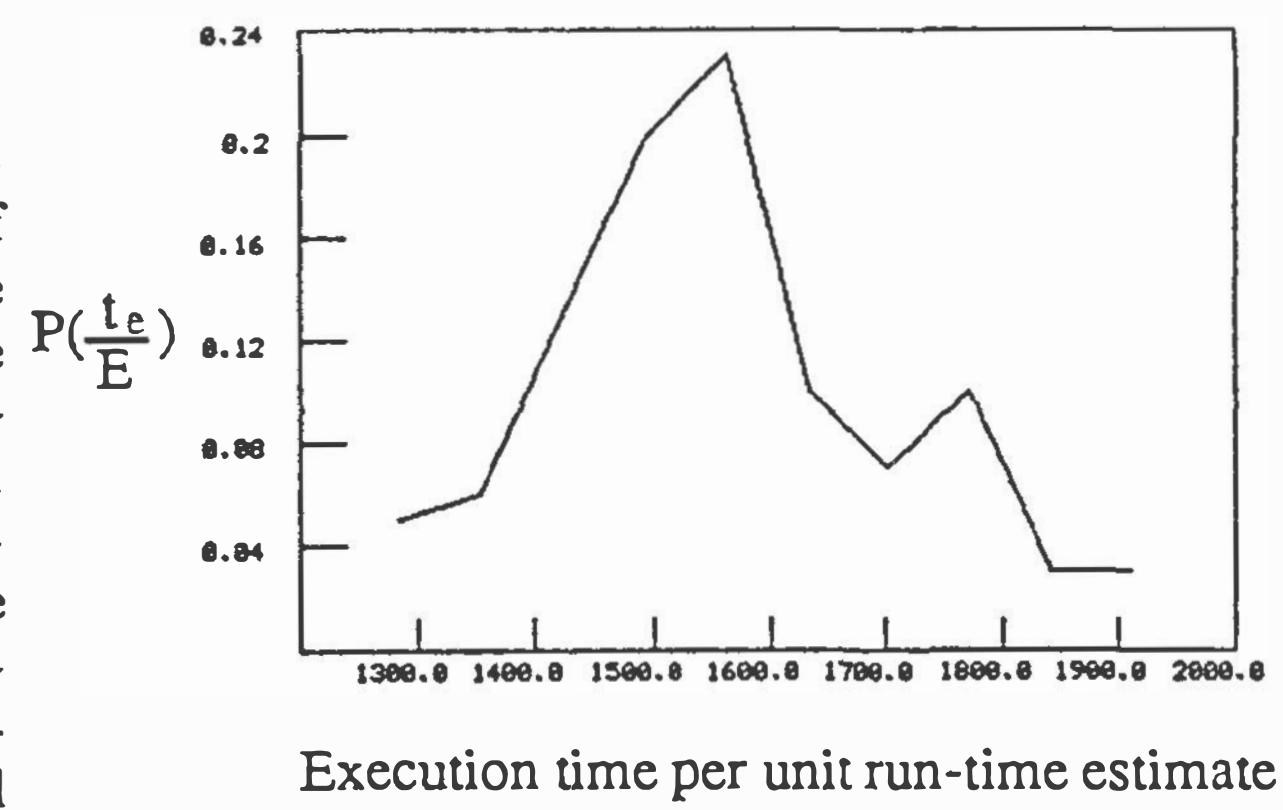

Figure 6: Probability distribution over time per operation for the clustering belief-network algorithm.

For a particular network, the benefits of additional reformulation time are uncertain because of the interplay between the K-search procedure and he specific interconnectedness and the state-space size for individual variables in a networks. Also, the expected incremental reduction in runtime is a function of how long the search has progressed. For the incremental algorithm, we therefore need to assess the probability distribution over the proportional reduction in $E_{tr+\Delta}$ from $E_{t_r}$ for various levels of $t_r$. Trajectories such as displayed in Figure 4 were analyzed to generate those probabilities. The actual data sets were collected for 0.5 and 0.25 second increments and applied in 0.5 second increments. Figure 5 displays one of these distributions. The graph shows the probability distribution over various levels of proportional decrease in estimated run time. The value at the left (zero) is the probability that the new reformulations searched in the next period will be no better than the current best. Positive probabilities to the right indicate the chances for improved execution times, given additional reformulation time. We found that the K-search technique performs so well, there are usually only minor gains to obtained from additional reformulation with this technique.

### 5.3.2 Time of Execution Given Estimate

The conditional independence assumption of Equation 5 allows us to assess a distribution over execution times, given the values reported by the estimator $E$, independent of the particular value $t_r$. We assessed this distribution by generating $E$ values for a number of random networks, and then timing the actual solution of each network. Since $E$ is based on total state space size, we divided actual execution time by $E$ to get a measure of execution time per unit run-time estimate used to generate $p(t_e|E_{t_r}, \xi)$. This distribution is shown in Figure 6.

The distributions do not reflect differences in possible configurations of evidence on the network. The estimator $E$ is based in the sum of the sizes of the state spaces for the cliques in the join tree for the constructed network. Since the problem we are analyzing consists of network construction, clique tree reformulation and a single inference cycle, the inference step we analyze is that of full propagation and initialization of the network, including the initial evidence vector. In analyzing clique tree formation for a network that would be applied to many possible evidence configurations, it would be necessary to examine the impact of different classes of evidence on the execution time.

## 5.4 Applying the Techniques

The criteria of Equation 6 has been implemented in a recent version of the IDEAL belief-network environment. Given a specification of a value function and a belief network, the system uses empirical data to determine, in real time, whether or not to continue with reformulation.

We performed an investigation of the value of metareasoning for optimizing the reformulation time. Because an analysis of the ideal reformulation of belief networks is sensitive to the efficiencies of the software and hardware, as well as to the formulation of the metareasoning model, it is important to consider details of the software and hardware. All experiments were run with IDEAL on a Symbolics 3645 Lisp Machine with 8 megabytes of physical memory.

The following experimental procedure was undertaken: A series of 30-node belief networks were constructed by a random belief-network generator in IDEAL. For each network and value-function pair, we applied (1) a *default* policy of halting reformulation after the first clique tree is identified by the K-search heuristic and (2) the incremental reformulation policy presented in Section 4, based on searching through a series of clique trees. After applying each technique, we executed an inference cycle (full propagation and marginalization of all nodes in the network), given evidence. The total time ($t_r + t_e$) was used to score the computational value of each trial based on the value function. This procedure was applied to a series of random belief networks for a given value function to assess the longterm performance of the default or incremental strategy. Given the metalevel model and the classes of probability distributions described above, we explored the relative efficacy of the default and incremental analyses for several value functions and parameterizations of these functions. These functions are shown in Figure 7.

Our analysis showed that the use of metareasoning to dynamically optimize the amount of time expended on reformulation frequently is more valuable than the static policy of halting reformulation after the first valid clique-tree is discovered. We found that the preferred approach, in terms of higher expected value over a number of trials, depended on the form of the value function and its specific parameters. The incremental metareasoning procedure continue to search if the benefit of finding a better clique formulation is high enough to justify the delay associated with continuing another time-increment of search. Since the K-search heuristic provides a very good initial clique formulation, (see Figures 5) incremental searching does not tend to provide a great deal of improvement in absolute terms. However when the costs of incremental delay are substantial, as in some of the quadratic value functions analyzed, the benefits of an improved solution can be substantial—even when these improvements are expected with low probability.

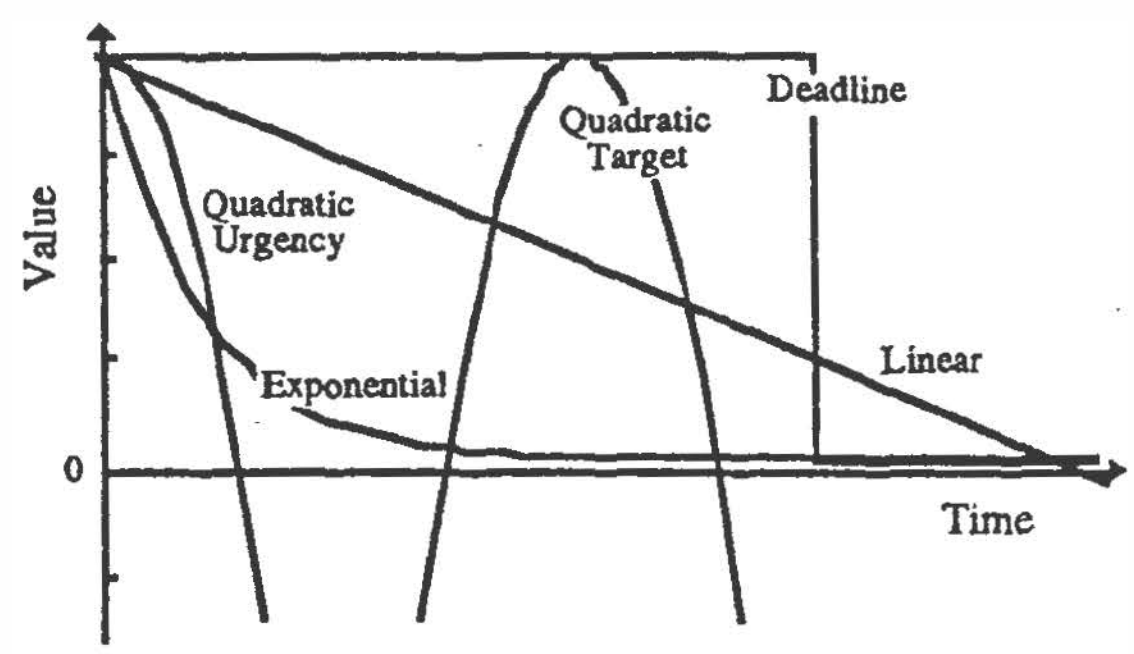

Figure 7: Prototypical value functions used in the incremental analyses.

For the linear and exponential forms with a slow decay of value with time, the incremental policies tend to behave like the default policy, as they stop searching for better cliques immediately after the first time increment. In these cases, the incremental policy was just marginally worse than the default policy.

For deadline models, we examined several variants by changing the severity of the deadline. We found that both policies performed equally well under a variety of deadlines, indicating that the ability to make the deadline was more dependent on the variability in the time required to perform inference on different networks (due to topology and state space size) than on differences in metalevel reasoning policy.

# 6 Summary

We described the metareasoning-partition problem and presented principles for calculating the ideal partition of resources under uncertainty for several prototypical classes of uncertainty and utility. We discussed the global optimization of the apportionment of resources for the case of a precursory reformulation where the reformulated instance is solved once. After, we introduced the incremental analyses for including evidence gleaned from observations about the progress of problem solving. Following the presenta-tion of our theoretical results, we discussed empirical study of the performance of a clique-tree reformulation strategy with these principles. We showed how an incremental reasoner can reason about the value of apportioning additional time to a search for optimal clusters in belief networks versus halting and solving the current best formulation. We found that the value of applying metalevel machinery to optimize the partition of resources for metareasoning is sensitive to the preference model, describing the value of a solution as a function of the delay needed for its computation. We hope that other investigators will find use in the principles we described for the ideal partition of resources for reformulation under uncertainty. In particular, the techniques hold the promise for helping us to optimally control the dynamic construction and solution of belief networks.